\documentclass[a4paper, 12pt]{article}
\usepackage[a4paper, margin=1in]{geometry}

\usepackage{amsmath}
\usepackage{graphicx}
\usepackage{amsfonts}
\usepackage{multirow}
\usepackage[numbers]{natbib}

\title{Interpretable Representation Learning from Videos using Nonlinear Priors \vspace{0.8cm}}
\author{
    \normalsize{\textbf{Marian Longa}} \\
    \normalsize{Visual Geometry Group} \\
    \normalsize{University of Oxford} \\
    \normalsize{\texttt{mlonga@robots.ox.ac.uk}} 
    \and
    \normalsize{\textbf{João F. Henriques}} \\
    \normalsize{Visual Geometry Group} \\
    \normalsize{University of Oxford} \\
    \normalsize{\texttt{joao@robots.ox.ac.uk}} 
}

\date{}
\begin{document}

\maketitle


\begin{abstract}
Learning interpretable representations of visual data is an important challenge, to make machines' decisions understandable to humans and to improve generalisation outside of the training distribution.
To this end, we propose a deep learning framework where one can specify nonlinear priors for videos (e.g. of Newtonian physics) that allow the model to learn interpretable latent variables and use these to generate videos of hypothetical scenarios not observed at training time.
We do this by extending the Variational Auto-Encoder (VAE) prior from a simple isotropic Gaussian to an arbitrary nonlinear temporal Additive Noise Model (ANM), which can describe a large number of processes (e.g. Newtonian physics).
We propose a novel linearization method that constructs a Gaussian Mixture Model (GMM) approximating the prior, and derive a numerically stable Monte Carlo estimate of the KL divergence between the posterior and prior GMMs.
We validate the method on different real-world physics videos including a pendulum, a mass on a spring, a falling object and a pulsar (rotating neutron star). We specify a physical prior for each experiment and show that the correct variables are learned.
Once a model is trained, we intervene on it to change different physical variables (such as oscillation amplitude or adding air drag) to generate physically correct videos of hypothetical scenarios that were not observed previously.
\end{abstract}

\section{Introduction}

Humans have an innate ability to observe the world and form a mental model of how things work -- if I release a hammer it will fall to the ground. Even more fascinating is the human ability to also reason about hypothetical scenarios which they may never have observed before -- what would happen if I dropped the same hammer while standing on Jupiter instead of on the Earth? In this example having a good mental model of how the position of a hammer depends on gravitational acceleration enables one to reason about the hammer's position on planets with different gravity, even though one has never actually performed such experiment there. In the causality literature this is referred to as counterfactual reasoning \cite{pearl_causality_2009,peters_elements_2017}, and it relies on having a Structural Equation Model describing the data that can be intervened on to obtain answers to counterfactuals, i.e. hypothetical questions.

There is a vast literature on traditional causality where many works aim to identify causal relations between variables that are provided by domain experts (such as economists or medical personnel) \cite{spirtes_algorithm_1991,chickering_optimal_2003}, however in many real world cases the variables are not specified a priori -- instead we want to learn them. Separately, in the field of intuitive physics researchers have been working on developing machine learning models that can understand physical scenarios \cite{lerer_blocks_2016,wu_galileo_2015,wu_physics101_2016,yi_clevrer_2020,watters_visintnet_2017,battaglia_interactionnetworks_2016,fragkiadaki_billiards_2015,mottaghi_newtonianscene_2016,wu_deanimation_2017,chang_neuralphysicsengine_2017,zheng_percepprednet_2018,deavila_diffphys_2018,janner_reasoning_2019,baradel_cophy_2020,smith_voe_2019,riochet_intphys_2018}, often with the aim to predict future behaviour of a system -- however, the works seldom focus on answering counterfactual questions in different situations.

In between the two, the field of causal representation learning \cite{scholkopf_towards_2021} has recently emerged
to train models to predict variables that are causally related, which can include physical variables.
This requires somewhat strong
assumptions, as it has been shown that learning a causal model purely from observational data is not possible \cite{locatello_challenging_2019}. Some works rely on the knowledge of interventions to infer the correct causal structure \cite{lippe_icitris_2022}, while other works make assumptions on the distribution of the latent variables and the causal mechanisms \cite{hyvarinen_tcl_2016,hyvarinen_td_2017,klindt_slowvae_2021,khemakhem_conditionalvae_2020,khemakhem_icebeem_2020,zimmermann_cl_2021,gresele_multiviewica_2020}.
Instead, in our work we assume a single variable and a known causal mechanism that is an additive noise model, which allows us to guarantee that the learned representation is identifiable and interpretable.

More specifically, if we consider the case of a hammer being dropped to the ground, we know that its vertical position depends quadratically on time and linearly on the gravitational acceleration ($y=\frac{1}{2} g t^2$, Newton's second law), which enables us to learn a variable corresponding to the object's position (assuming there are no confounders). We can then intervene on this model by changing the gravitational acceleration to reason about the object falling on different planets and visualise the resulting videos. In this paper we learn such models for different physical scenarios, including a falling object, an object oscillating on a spring, and object swinging on a pendulum, and a pulsating Crab pulsar. For each video of a real-world physics experiment we specify a prior (physical equation) that can be intervened to generate an imagined, previously-unobserved case of the same scenario but with different physical properties, specifically an object falling on Jupiter, a pendulum with added air resistance, a mass on a spring oscillating with a different mass and frequency, and a pulsar pulsating at a different frequency.

To learn the physical variables we use a Variational Auto-Encoder (VAE) \cite{kingma_auto-encoding_2014}. However, instead of a standard isotropic Gaussian prior, we use an Additive Noise Model (ANM) which is a general model of the form $y=f(x)+n$ (where $n$ is Gaussian noise), that can describe a wide range of physical systems -- for example, for a video of a mass oscillating on a spring, we can specify the prior as $y(t)=A\cos(\omega t)+n$.
Note that we assume a single variable and a known physical mechanism, which is crucial to obtain identifiability.
We then devise a method to approximate this prior as a Gaussian Mixture Model (GMM) and derive an expression for the KL divergence estimate to be able to optimise the data log-likelihood of the VAE. Once we train the model using a given ANM prior we can then intervene on the prior and decode a video corresponding to an imagined scenario given by the new model -- for example, $y(t)=\frac{1}{2}A\cos(2\omega t)-\frac{1}{2}+n$. In this way we have developed a method for counterfactual reasoning in videos. 

Concretely, our contributions in this paper are:
\begin{enumerate}
    \item A novel interpretable representation learning framework based on a VAE with a nonlinear Additive Noise Model prior, for learning temporal variables from video, and the ability to intervene on the model to produce realistic counterfactual videos.
    \item A novel method for approximation of the prior density via local linearizations of the prior, automatically decomposing it into a mixture of non-isotropic Gaussians.
    \item A numerically stable and highly parallelizable KL divergence estimate between prior and posterior distributions.
    \item Experiments on 4 real videos, showing that the method learns the correct latent variables and is able to generate realistic videos outside of the training distribution.
\end{enumerate}

\section{Related Work}\label{sec:related}
\paragraph{Causality.} In the traditional causal inference literature the aim is mostly to infer causal relationships between variables provided by domain experts \cite{spirtes_algorithm_1991,chickering_optimal_2003} (for an overview of methods see \cite{peters_elements_2017}). More recently the field of causal representation learning has emerged, which aims to learn a causal representation of data that includes the mechanisms relating the variables as well as the variables themselves (for overview see \cite{scholkopf_towards_2021}). It has been shown that one cannot learn both the true latent variables and mechanisms relating them at the same time without making additional assumptions on the data and the model \cite{locatello_challenging_2019}, thus different causal representation learning works make different assumptions to achieve identifiability \cite{hyvarinen_tcl_2016,hyvarinen_td_2017,klindt_slowvae_2021,khemakhem_conditionalvae_2020,khemakhem_icebeem_2020,zimmermann_cl_2021,gresele_multiviewica_2020} (for an overview of identifiability assumptions for different works see \cite{ahuja_properties_2022}). In this work we achieve identifiability by providing a physical mechanism prior describing a variable's evolution of time. Unlike other works that employ the knowledge of mechanisms, we focus on applications to high-dimensional unstructured data in the form of real videos and specifically focus on physical mechanisms.

\paragraph{Intuitive physics.} In the intuitive physics literature the aim is to design systems that mimic the human ability to use ``common sense'' to understand physical phenomena. Broadly, Duan et al. \cite{duan_intuitivephysicssurvey_2022} categorise intuitive physics works by the type of the physical reasoning task each work aims to solve: prediction of the outcome of interactions between physical objects \cite{lerer_blocks_2016,wu_galileo_2015,wu_physics101_2016,yi_clevrer_2020}, prediction of the dynamics (trajectories) of objects \cite{lerer_blocks_2016,watters_visintnet_2017,wu_physics101_2016,battaglia_interactionnetworks_2016,fragkiadaki_billiards_2015}, inference of physical properties of objects \cite{mottaghi_newtonianscene_2016,wu_deanimation_2017,wu_galileo_2015,chang_neuralphysicsengine_2017,fragkiadaki_billiards_2015,zheng_percepprednet_2018,deavila_diffphys_2018}, generating physically plausible future video frames \cite{janner_reasoning_2019,watters_visintnet_2017,zheng_percepprednet_2018,deavila_diffphys_2018,baradel_cophy_2020}, predicting whether a scenario is physically possible or not (``violation of expectation'') \cite{smith_voe_2019,riochet_intphys_2018}, curriculum learning, counterfactual prediction and physical equation prediction \cite{baradel_cophy_2020,yi_clevrer_2020}. Our work falls most closely in the category dealing with inference of physical properties, as we aim to learn physical variables such as object positions and angles, while also overlapping with the counterfactual prediction and future frame prediction. Unlike existing works, our work is primarily grounded in causality as opposed to the empirical focus in other works, including specialized object detection modules, physics simulators, and deep learning methods which sometimes lack strong theoretical guarantees. 

\paragraph{VAE priors.} Another area of research is developing VAEs with differently structured priors where the aim is usually to design a prior that achieves good image reconstructions on a standard set of image benchmarks. For example, the VAE \cite{kingma_auto-encoding_2014}, $\beta$-VAE \cite{higgins_betavae_2017} and FactorVAE \cite{kim_disentangling_2018} aim to learn independent latent variables, while Ladder-VAE \cite{sonderby_ladder_2016} conditions each variable on the preceding one and NVAE \cite{vahdat_nvae_2020} and VDVAE \cite{child_vdvae_2021} condition each variable on all the previous variables. While we also use a VAE with a structured prior, our prior is given by an ANM that comes from our desire to model physical processes. Also, instead of aiming to achieve the best image reconstructions, our primary focus is to learn physical latent variables as part of a structural equation model that can be intervened on in order to produce counterfactual videos.

\section{Method} \label{sec:method}

We will now describe our main proposal, which extends VAEs with a causal prior for temporal data.

\paragraph{Optimisation Objective.} We start from a standard VAE \cite{kingma_auto-encoding_2014}, and consider the case when all input samples correspond to different temporal samples from the same video.
The optimisation objective is to maximise the evidence lower bound \cite{kingma_auto-encoding_2014}, $\mathrm{ELBO} = \mathbb{E}_{p(x)}[\mathbb{E}_{q(z|x)} \ln p(x|z) - D_{KL}(q(z|x) || p(z))]$.
%
%
The first term corresponds to a reconstruction error, given by the $L_2$ loss between the original image $x_t$ and its reconstruction $\hat{x}_t$. 
The second term corresponds to the Kullback-Leibler (KL) divergence between the posterior and the prior distributions. We compute each as follows.

\begin{figure}[t]
\begin{center}
   \includegraphics[width=0.49\linewidth]{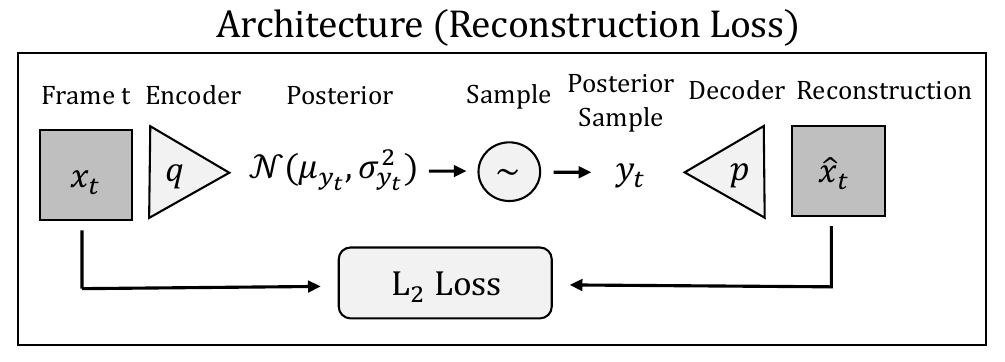}
   \includegraphics[width=0.49\linewidth]{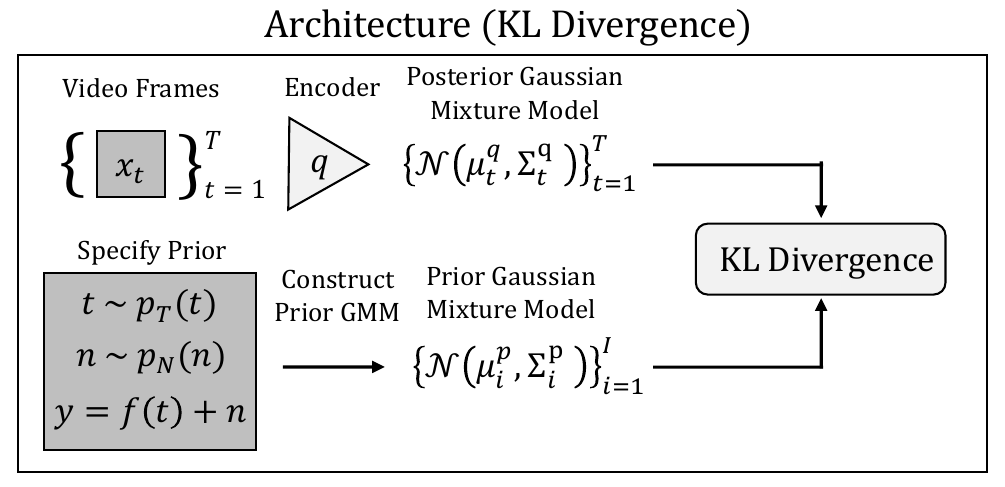}
\end{center}
\vspace{-5mm}
   \caption{Left: architecture to compute the reconstruction loss. Right: architecture to compute the KL divergence loss. See sec. \ref{sec:method} for details.}
\label{fig:architecture}
\end{figure}

\paragraph{Posterior GMM.} The posterior distribution consists of a mixture of $T$ Gaussians, each corresponding to a single frame of video (sample). Each frame $x_t$ is encoded independently by an encoder $q$ to obtain the parameters $\mu_{y_t}$ and $\sigma_{y_t}$ which are concatenated with time $t$ to obtain the posterior Gaussian 
\begin{align} \label{eq:posterior-gmm}
    \mathcal{N}(\mu_t^q,\Sigma_t^q) = \mathcal{N}\left(\begin{bmatrix}t \\ \mu_{y_t}\end{bmatrix},\begin{bmatrix}\epsilon^2 & 0 \\ 0 & \sigma_{y_t}^2\end{bmatrix}\right)
\end{align}
In all experiments we set $\epsilon = 0.01$.
See fig. \ref{fig:architecture} (top branch of the right panel) for an illustration. To compute the reconstruction loss (fig. \ref{fig:architecture}, left panel) we only pass the second component of the sampled posterior to the decoder, namely $y_t \sim \mathcal{N}(\mu_{y_t}, \sigma_{y_t}^2)$, while discarding the first component $t \sim \mathcal{N}(t, \epsilon^2)$, in order to prevent the decoder solely relying on the time variable and ignoring the $y_t$ variable.



\paragraph{Prior GMM.} First, we specify the prior by choosing the time distribution $p_T(t)$, the noise distribution $p_N(n)$, and the mechanism $y=f(t)+n$, based on our knowledge of the physical process in a given video (which must be given). The idea is then to sample enough time instants $t_0 \sim p_T(t_0)$ and noise realizations $n_0 \sim p_N(n_0)$ to construct a Gaussian mixture model from the prior, where each Gaussian has mean $(t, y)$ with $t=t_0$ and $y=f(t_0)+n_0$, and with a covariance that follows the curvature of $f(t)$ at $t_0$.

Formally, we model each time as a Gaussian\footnote{The time $t$ is modelled as a Gaussian whose mean is set to the true time of the corresponding image frame and whose variance models the uncertainty in temporal sampling, which is related to the time difference between two consecutive image frames.} $t \sim \mathcal{N}(t_0, \sigma_t^2)$ centered around $t_0 \sim p_T(t_0)$ (e.g. $t_0 \sim U(0,1)$) with variance $\sigma_t^2$ set to a fixed value (here $\sigma_t=\epsilon=0.01$). Next, because a non-linear function of a Gaussian variable is no longer Gaussian, we approximate $y=f(t)+n$ linearly around $t_0$ using its first order Taylor expansion as
\begin{align} \label{eq:taylor}
    y \approx f(t_0) + \frac{\partial f}{\partial t}\Big|_{t_0}(t-t_0)+n.
\end{align}
Using eq. \ref{eq:taylor} the joint distribution of $t$ and $y$ can now be written as
\begin{align} \label{eq:joint-t-y}
    \begin{bmatrix}t \\ y\end{bmatrix} = \begin{bmatrix}1 \\ \frac{\partial f}{\partial t}\Big|_{t_0}\end{bmatrix} t + \begin{bmatrix}0 \\ 1\end{bmatrix} n + \begin{bmatrix}0 \\ f(t_0) - \frac{\partial f}{\partial t}\Big|_{t_0}t_0\end{bmatrix}.
\end{align}
%
Using the formula for linear combinations of Gaussian variables (\cite{bishop_pattern_2006} Sec. 8.1.4) the joint probability in eq. \ref{eq:joint-t-y} can be expressed as $(t,y) \sim \mathcal{N}(\mu^p, \Sigma^p)$ where
\begin{align}
    \mu^p = \begin{bmatrix}1 \\ \frac{\partial f}{\partial t}\Big|_{t_0}\end{bmatrix} t_0 + \begin{bmatrix}0 \\ 1\end{bmatrix} n_0 + \begin{bmatrix}0 \\ f(t_0) - \frac{\partial f}{\partial t}\Big|_{t_0}t_0\end{bmatrix},\
    \Sigma^p = \begin{bmatrix}1 \\ \frac{\partial f}{\partial t}\Big|_{t_0}\end{bmatrix} \sigma_t^2 \begin{bmatrix}1 \\ \frac{\partial f}{\partial t}\Big|_{t_0}\end{bmatrix}^T + \begin{bmatrix}0 \\ 1\end{bmatrix} \sigma_n^2 \begin{bmatrix}0 \\ 1\end{bmatrix}^T, \label{eq:mu-sigma-p}
\end{align}
where we have used $t \sim \mathcal{N}(t_0, \sigma_t^2)$ and $n \sim \mathcal{N}(n_0, \sigma_n^2)$. Finally, simplifying equations \ref{eq:mu-sigma-p} yields the prior
\begin{align} \label{eq:prior-gmm}
    \begin{bmatrix}t \\ y\end{bmatrix} \sim  \mathcal{N}\left(\begin{bmatrix}t_0 \\ f(t_0)+n_0\end{bmatrix}, \begin{bmatrix}\sigma_t^2 & \frac{\partial f}{\partial t}\Big|_{t_0} \sigma_t^2 \\ \frac{\partial f}{\partial t}\Big|_{t_0} \sigma_t^2 & \left(\frac{\partial f}{\partial t}\Big|_{t_0}\right)^2 \sigma_t^2 + \sigma_n^2\end{bmatrix}\right).
\end{align}
The prior GMM is thus constructed as a mixture of the Gaussians specified in eq. \ref{eq:prior-gmm} with equal weight. See fig. \ref{fig:architecture} (bottom branch of the right panel) for an illustration.


\paragraph{KL Divergence.} Because the prior and posterior are both GMMs (as opposed to single Gaussians as in a standard VAE), the KL divergence can no longer be computed in closed form. We can nevertheless approximate the KL divergence using Monte Carlo sampling \cite{bishop_pattern_2006} (chapter 11) as
\begin{align} \label{eq:kl-monte-carlo}
    D_{KL}(q||p) \approx \frac{1}{I}\sum_{i=1}^{I}[\ln q(x_i) - \ln p(x_i)],\ x_i \sim q(x)
\end{align}
In words, we first sample the posterior GMM to obtain $x_i$, and then we evaluate this sample in the GMM prior $p(x)$ and posterior $q(x)$ to obtain the estimate, and we repeat this $I$ times. However, naively evaluating $p$ and $q$ in eq. \ref{eq:kl-monte-carlo} introduces many numerical instabilities, and the expression is not immediately parallelizable over both the posterior samples ($x_i$) and the prior and posterior Gaussian mixtures ($p$, $q$) at the same time. To address both issues we derive a numerically stable and parallelizable form of eq. \ref{eq:kl-monte-carlo}, by deriving the form of $\ln p(x_i)$ and $\ln q(x_i)$ directly in terms of the parameters of individual Gaussians in each GMM. Substituting the form of a GMM for $p(x)$ and evaluating the probability density function (PDF) of each Gaussian yields
\begin{align} \label{eq:kl-monte-carlo-2}
    \ln p(x_i) = \ln \sum_{j=1}^{N}\frac{1}{N}\mathcal{N}(x_i; \mu_j, \Sigma_j)
    = \ln \sum_{j=1}^N\frac{1}{N}(2\pi)^{-\frac{k}{2}} \det(\Sigma_j)^{-\frac{1}{2}}e^{-\frac{1}{2}(x_i-\mu_j)^T\Sigma_j^{-1}(x_i-\mu_j)}.
\end{align}
Simplifying eq. \ref{eq:kl-monte-carlo-2} and factoring terms to merge numerically unstable operations (such as independently-computed determinants and logarithms) yields
\begin{align} \label{eq:kl-monte-carlo-3}
\ln p(x_{i})=-\ln N-\frac{k}{2}\ln2\pi+\overset{\mathtt{logsumexp}}{\overbrace{\ln\sum_{j=1}^{N}\exp}}\Big[
-\frac{1}{2}\underset{\mathtt{logdet}}{\underbrace{\ln(\det(}}\Sigma_{j}))-\frac{1}{2}(x_{i}-\mu_{j})^{T}\Sigma_{j}^{-1}(x_{i}-\mu_{j})\Big].
\end{align}
Eq. \ref{eq:kl-monte-carlo-3} now only contains numerically stable operations, namely linear combination of terms, {\tt logsumexp} and {\tt logdet}. It can also be efficiently computed for an arbitrary number of posterior samples (indexed by $i$) and prior GMM Gaussians (indexed by $j$) in parallel, unlike eq. \ref{eq:kl-monte-carlo}. An identical expression to eq. \ref{eq:kl-monte-carlo-3} can be derived for the posterior $q(x_i)$ instead of the prior $p(x_i)$ and subtracting the two according to eq. \ref{eq:kl-monte-carlo} yields the KL divergence estimate.

\begin{table}[t]
\begin{center}
\small
\begin{tabular}{|c|c|c|c|}
\hline
\textbf{Experiment} & \textbf{Time Prior} & \textbf{Mechanism Prior} & \textbf{Mechanism Intervention} \\
\hline\hline
Mass on a Spring & $t \sim U(0,2\pi)$ & $y = \cos(t)$ & $y = \frac{1}{3} \cos(2t) -\frac{2}{3}$ \\
\hline
Pendulum & $t \sim U(0,2\pi)$ & $\theta = \cos(t)$ & $\theta = e^{-t/2\pi}\cos(t)$ \\
\hline
Falling Object & $t \sim U(0,1)$ & $y = 1 - t^2$ & $y = \max(0, 1 - 2.53t^2)$ \\
\hline
Pulsar & $t \sim U(0,1)$ & 
\begin{tabular}{@{}c@{}}$I = 0.125\ \mathcal{N}(t; 0.055, 0.05^2)$ \\ $+ 0.085\ \mathcal{N}(t; 0.5, 0.08^2)$\end{tabular} & 
\begin{tabular}{@{}c@{}}$I = 0.0193\ \mathcal{N}(t; 0.195, 0.007^2)$ \\ $+ 0.006\ \mathcal{N}(t; 0.6, 0.008^2)$\end{tabular} \\
\hline
\end{tabular}
\end{center}
\vspace{-2mm}
\caption{Summary of the prior time distribution, the prior mechanism equation and the intervened mechanism equation for the experiments with mass on a spring, pendulum, falling object, and Crab pulsar.}
\label{tab:results}
\end{table}

\section{Experiments} \label{sec:experiments}
\begin{figure}[t]
\begin{tabular}{cccc}
\includegraphics[width=0.22\linewidth]{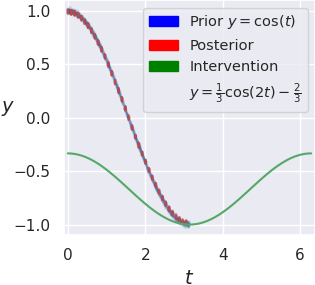} &
\includegraphics[width=0.22\linewidth]{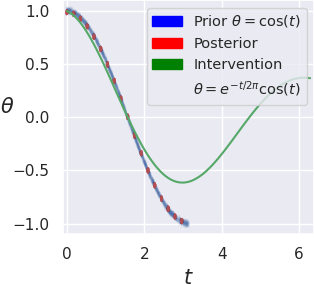} &
\includegraphics[width=0.22\linewidth]{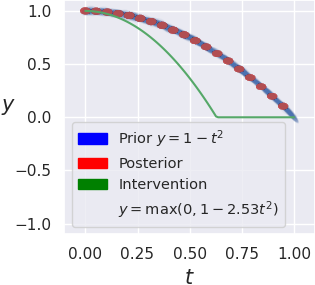} &
\includegraphics[width=0.22\linewidth]{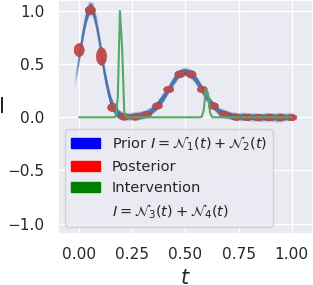}\\
(a) Mass on a spring & (b) Pendulum & (c) Falling object & (d) Pulsar
\end{tabular}
\caption{Latent space visualisation showing the prior generated using the linearization method (blue), posterior generated by encoding many images from the dataset (red), and the intervened prior (green), shown for experiments (a-d).}
\label{fig:latents}
\end{figure}

In this section we demonstrate our method on 4 physics experiments: a mass oscillating on a spring, a mass swinging on a pendulum, an object falling from a height and a pulsar pulsating over time. Even though our method only deals with a single variable with a known mechanism, these experiments demonstrate its practical applications. Each experiment has high-dimensional observations, in the form of a raw RGB video. For each experiment we specify a physics prior for the variable that we want the model to learn, and once the model is learned the prior is intervened on to generate a never-before-seen video of a modified experiment -- for example seeing how a pendulum would behave with air resistance or seeing how an object would fall on Jupiter instead of on Earth. This method enables one to visualise many hypothetical scenarios (counterfactuals) that have never been seen before at training time but still result in realistic, physically plausible videos. The equations of the prior and interventions are summarised in table \ref{tab:results} and are visualised in fig. \ref{fig:latents}, and the corresponding prior and intervened videos are shown in 
fig. \ref{fig:video}.

\paragraph{Implementation.} We implement the method described in sec. \ref{sec:method} using the architecture in fig. \ref{fig:architecture}. The encoder $q$ and decoder $p$ are both 4-layer multilayer perceptrons (MLPs) with 128 hidden units and with ReLU activations between each layer. We optimise the ELBO term (sec. \ref{sec:method}, Optimisation Objective) using the Adam optimiser \cite{kingma_adam_2015} with learning rate $3\cdot 10^{-4}$ and batch size 100, until convergence. Image dimensions are $3 \times 100 \times 226$ px for the Spring experiment, $3 \times 100 \times 108$ px for the Pendulum experiment, $3 \times 100 \times 205$ px for the Falling Object experiment, and $3 \times 77 \times 78$ px for the Pulsar experiment. 

\paragraph{Baselines.} In each experiment we train two baseline VAEs with a standard $\mathcal{N}(0,1)$ prior
\cite{kingma_auto-encoding_2014}. The decoder of one baseline has the video time appended to its input, while the other does not (Temporal VAE and Standard VAE in table \ref{tab:baselines}). All other training details are identical to our model.

\subsection{Mass on a Spring} \label{sec:spring}
\begin{figure}[t]
\setlength{\tabcolsep}{1pt} 
\begin{tabular}{cccc}
\includegraphics[width=3.6cm]{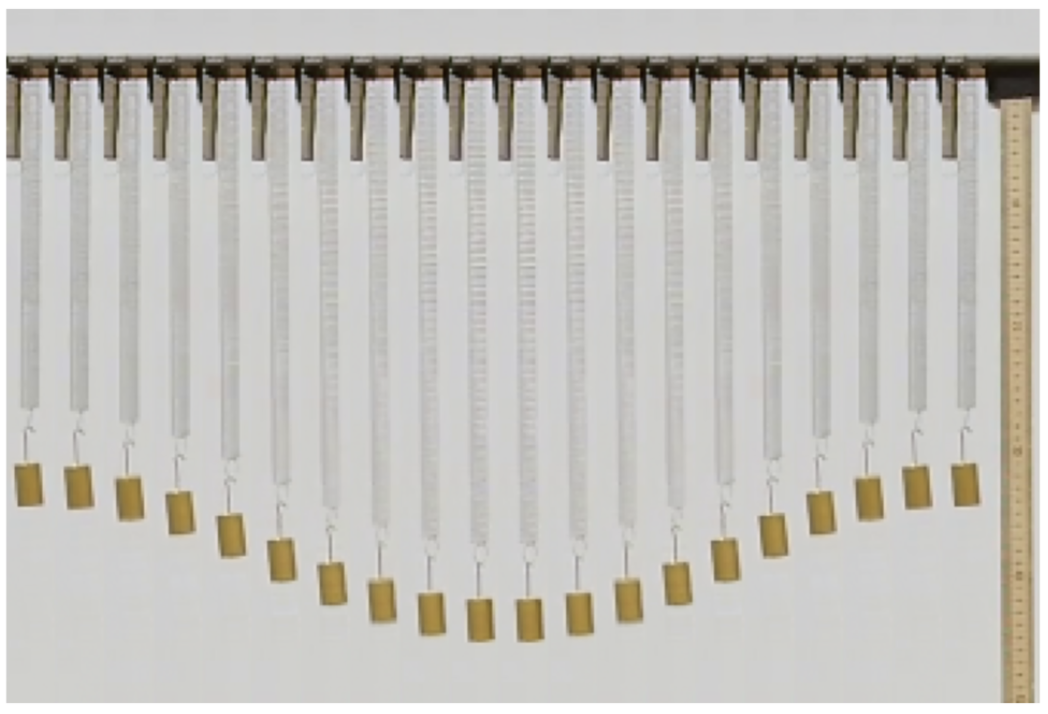} &
\raisebox{2.15cm}{\multirow{2}{*}{\includegraphics[width=2.05cm]{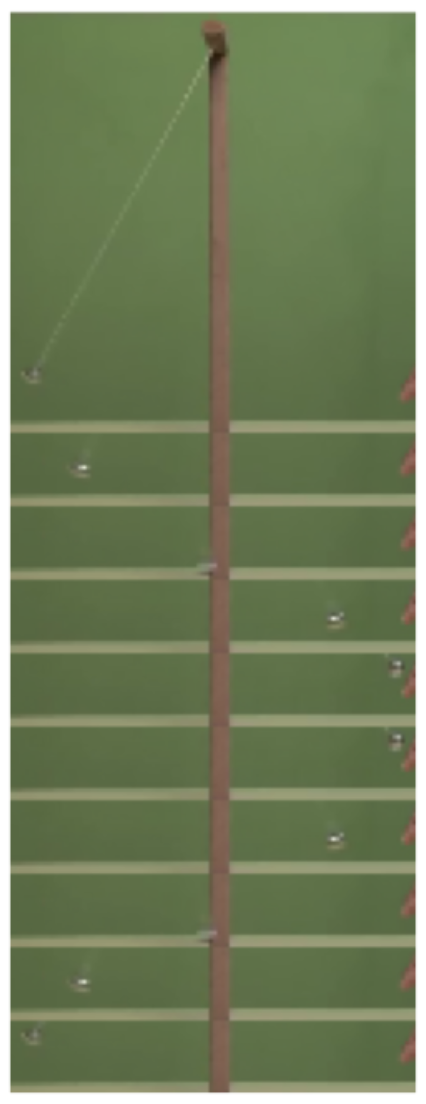}\includegraphics[width=2.1cm]{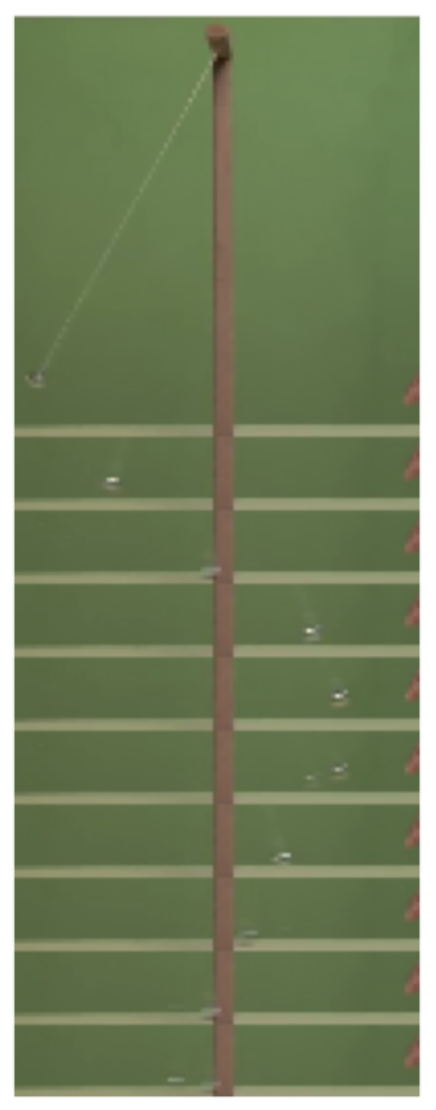}}} &
\includegraphics[width=4.7cm]{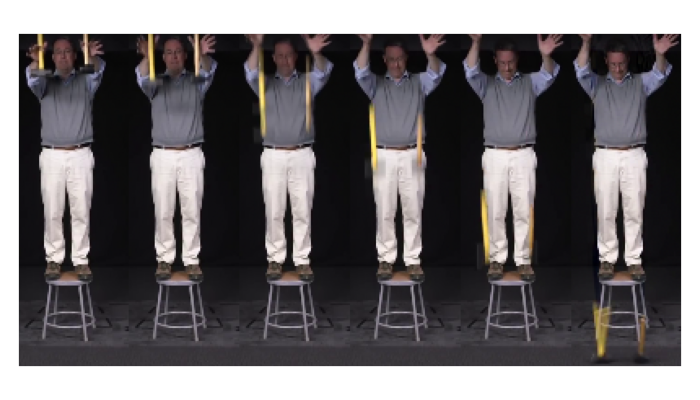} &
\includegraphics[width=3cm]{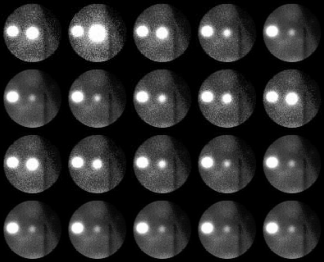} \\
\includegraphics[width=3.6cm]{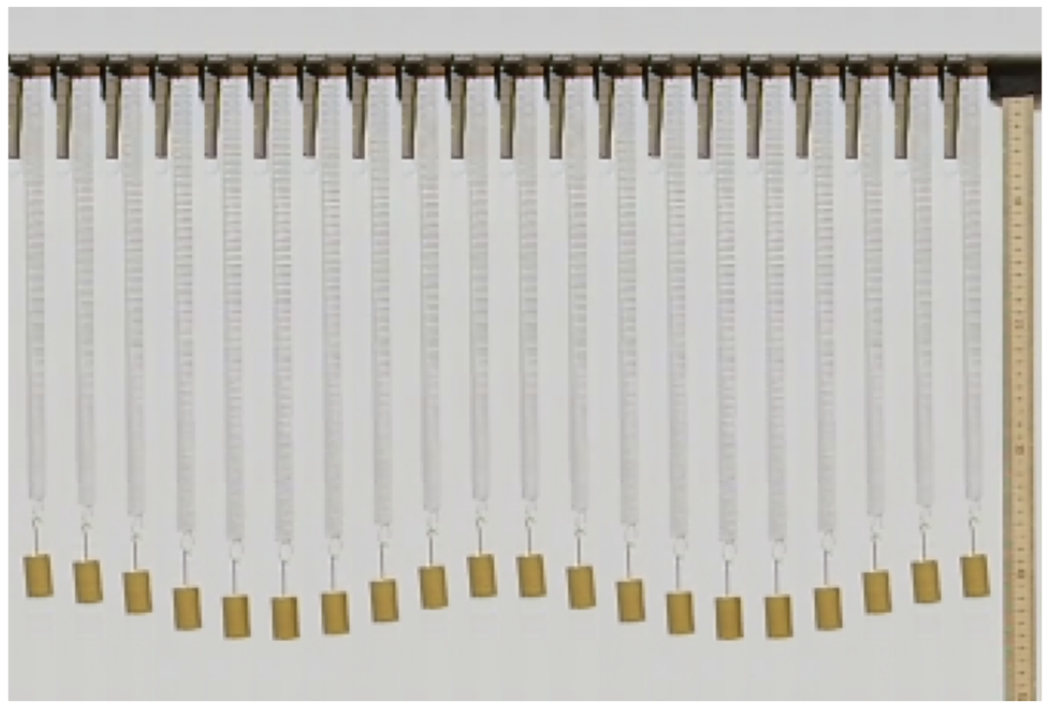} &
&
\includegraphics[width=4.7cm]{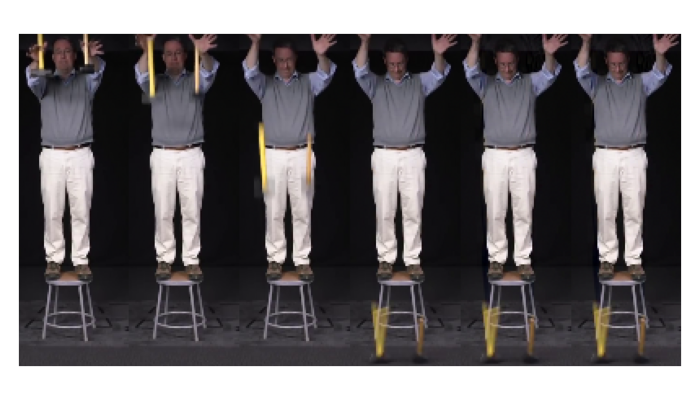} &
\includegraphics[width=3cm]{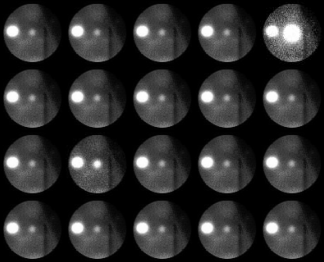} \\
(a) Mass on a spring & (b) Pendulum & (c) Falling object & (d) Pulsar
\end{tabular}
\caption{Original and counterfactual videos for experiments (a-d) obtained by decoding the original and intervened priors.}
\label{fig:video}
\end{figure}

In this video (fig. \ref{fig:video} (a), top) a mass is hung from a spring and is left to oscillate with a certain oscillation amplitude and frequency around a certain height, with negligible air resistance. Assuming mass $m$, spring constant $k$, acceleration due to gravity $g$ and vertical mass position $y$, the equation of motion for the mass can be derived by solving Newton's second law for $y$, as $m \frac{\partial^2 y}{\partial t^2} = -ky +mg$.
%
%
Assuming the mass starts at its highest position, the solution to this equation is given by $y(t)=A \cos(\omega t)+h,\ \omega=\sqrt{\frac{k}{m}},\ h=\frac{mg}{k}$.
%
%
For simplicity we set $A=1$, $\omega=1$ and $h=0$ to obtain the prior $y(t)=\cos(t)$.

\paragraph{Quantitative results.}
We plot the prior (blue) and posterior (red) in fig. \ref{fig:latents} (a) and note that the posterior matches the prior very closely. In table \ref{tab:baselines} (first row) we show that our method achieves the lowest mean squared error (MSE) between the predicted and the estimated ground truth position of the object (after normalising the scale and offset of each distribution) among all baselines by 3 to 4 orders of magnitude, showing that our model has learned the correct physical variables. Finally, in table \ref{tab:baselines} (first row) we show that our model achieves a similarly good reconstruction accuracy (Rec. Acc.) as the temporal VAE baseline while outperforming the standard VAE, where the (seemingly small) $0.5\%$ difference in accuracy means that the standard VAE only reconstructed the average frame across the whole video, while our method and the temporal VAE correctly reconstructed each video frame.

\paragraph{Intervention results.}
As an example of counterfactual reasoning, consider that we want to generate a video of an experiment where the amplitude of oscillation is a third of the original one ($A_{int}=1/3 A$), the mass is four times lighter ($m_{int}=m/4 \implies \omega_{int}=2\omega$) and the oscillation point is two thirds below the original one ($h_{int}=h-2/3$). Thus, for such an intervention we can modify the prior as $y_{int}(t)=1/3 \cos(t) -2/3$ (fig. \ref{fig:latents} (a), green) and decode it to obtain the corresponding video (fig. \ref{fig:video} (a), bottom). The counterfactual video shows a realistic demonstration of a mass oscillating around a lower point with a smaller amplitude and a higher frequency.

\subsection{Pendulum} \label{sec:pendulum}
In this video (fig. \ref{fig:video} (b), left) a mass is suspended from a string and is left to oscillate from left to right to form a pendulum with negligible air resistance. Assuming mass $m$, string length $L$, string angle from vertical $\theta$, coefficient $c$ related to air resistance (for generality) and acceleration due to gravity $g$, the equation of motion for the mass can be derived by solving Newton's second law, as $mL\frac{\partial^2 \theta}{\partial t^2}=-mg\sin(\theta)+cL\frac{\partial \theta}{\partial t}$.
%
%
If the mass is released from angle $\theta_0$ this equation is solved by $\theta(t)=\theta_0 e^{-\gamma t}\cos(\omega t),\ \omega=\sqrt{\frac{g}{L}},\ \gamma=\frac{c}{2m}$.
%
%
For simplicity we set $\theta_0=1$, $\omega=1$ and assume no air drag by setting $\gamma=0$ to obtain the prior $\theta(t)=\cos(t)$.
%

\paragraph{Quantitative results.}
We plot the prior (blue) and posterior (red) in fig. \ref{fig:latents} (b) and note the posterior matches the prior very closely. Similar to the spring experiment, table \ref{tab:baselines} (second row) shows that our method achieves the lowest MSE among all baselines and a reconstruction accuracy similarly good to the temporal VAE, demonstrating that our method has learned the correct physical variable (pendulum angle) while successfully reconstructing the images unlike the baselines.

\paragraph{Intervention results.}
Now if we want to simulate a scenario where the mass is slowed down by air drag we can for example set $\gamma=1/2\pi$ in the Newton's law solution to obtain the intervened prior $\theta_{int}(t)=e^{-t/2\pi}\cos(t)$ (fig. \ref{fig:latents} (b), green). Decoding latent variables generated with this intervened prior results in a novel video where the mass slows down over time (fig. \ref{fig:video} (b), right). By changing the parameters in the Newton's law solution we could also obtain counterfactual videos of a mass oscillating with different strengths of air resistance (by intervening on $\gamma$), on different planets (by intervening on $g$), or with different amplitudes (by intervening on $\theta_0$).

\subsection{Falling Object} \label{sec:fall}
In this experiment an object is dropped from rest at an initial height and is left to accelerate and eventually hit the ground (fig. \ref{fig:video} (c), top). Assuming mass $m$, acceleration due to gravity $g$ and assuming the object is released from rest, the equation of motion is determined by solving Newton's second law, as $m\frac{\partial^2 y}{\partial t^2} = -mg$.
%
%
If the object is released from height $y_0$, this equation is solved by $y=y_0-\frac{1}{2}gt^2$.
%
%
For simplicity we set $y_0=1$ and $g=2$ to obtain the prior $y=1-t^2$.

\paragraph{Quantitative results.}
Fig \ref{fig:latents} (c) shows the quadratic prior (blue) matches the posterior (red). Table \ref{tab:baselines} (third row) shows that our method achieves the lowest MSE among all baselines, and a reconstruction accuracy comparable to the temporal VAE baseline.

\paragraph{Intervention results.}
Now if we want to simulate a scenario where the object is falling on the surface of Jupiter instead of on Earth's surface where $g_{\mathrm{Jupiter}}=2.53g_{\mathrm{Earth}}$, we can modify the prior as $y_{int}=1-2.53t^2$ (fig. \ref{fig:latents} (c), green) and decode it to obtain the corresponding video (fig. \ref{fig:video} (c), bottom) where the object now falls much faster. By intervening on the Newton's law solution we could also simulate many other hypothetical scenarios including throwing the object up from the ground with a given initial speed and waiting for it to come back down, or seeing how adding the effects of air resistance would change the object's trajectory, or making the object bounce off when it hits the ground and after a few bounces slowly come to rest.
For additional results see the appendix.

\begin{table}[t]
\begin{center}
\small
\begin{tabular}{|l|cc|cc|cc|}
\hline
& \multicolumn{2}{|c|}{\textbf{Standard VAE}} & \multicolumn{2}{|c|}{\textbf{Temporal VAE}} & \multicolumn{2}{|c|}{\textbf{Proposed Method}} \\
\hline
\textbf{Experiment} & Lat. MSE & Rec. Acc. & Lat. MSE & Rec. Acc. & Lat. MSE & Rec. Acc. \\
\hline\hline
Mass on a Spring & $1.96$ & $99.41\%$ & $2.00$ & $\mathbf{99.95\%}$ & $\mathbf{2.7 \cdot 10^{-4}}$ & $\mathbf{99.95\%}$ \\
\hline
Pendulum & $1.99$ & $99.69\%$ & $1.98$ & $99.95\%$ & $\mathbf{2.5 \cdot 10^{-4}}$ & $\mathbf{99.99\%}$ \\
\hline
Falling Object & $2.03$ & $98.33\%$ & $2.00$ & $\mathbf{99.96\%}$ & $\mathbf{2.0 \cdot 10^{-3}}$ & $99.85\%$ \\
\hline
Pulsar & $1.99$ & $95.82\%$ & $1.97$ & $\mathbf{99.98\%}$ & $\mathbf{7.0 \cdot 10^{-2}}$ & $97.56\%$ \\
\hline
\end{tabular}
\end{center}
\vspace{-1mm}
\caption{Mean squared error (Lat. MSE, lower is better) of the predicted posterior latent variables w.r.t. estimated ground truth physical variables, and reconstruction accuracy (Rec. Acc., higher is better) of the decoded video frames w.r.t. input video frames. Results are shown for the proposed method, a standard VAE and a temporally conditioned VAE.}
\label{tab:baselines}
\end{table}

\subsection{Pulsar} \label{sec:pulsar}
\begin{figure}[t]
\begin{center}
   \includegraphics[width=0.49\linewidth]{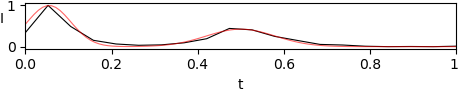}
   \includegraphics[width=0.49\linewidth]{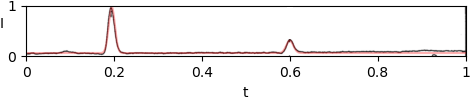}
\end{center}
    \vspace{-5mm}
   \caption{Crab pulsar intensity data (black) and a double Gaussian mixture fit (red) at the original UVB frequency (left) and at the intervention frequency 1418 MHz (right).}
\label{fig:intensity-pulsar}
\end{figure}

In this experiment we deal with image data from a pulsar, which is a rotating neutron star that emits periodic pulses of electromagnetic radiation. Here we use the image data of the Crab pulsar recorded at different phases in the UVB spectrum \cite{aaron_pulsaruvb_2000} (fig. \ref{fig:video} (d), top; based on fig. 2 of the cited paper). For each pulsar image we compute its intensity as the average of the pixel values across the image (fig. \ref{fig:intensity-pulsar}, left, black), and manually fit it with a Gaussian mixture model consisting of two Gaussians (fig. \ref{fig:intensity-pulsar}, left, red), corresponding to the two peaks in its emission profile. The prior intensity is thus given by $I(t)=A_1\mathcal{N}(t;\mu_1,\sigma_1^2)+A_2\mathcal{N}(t;\mu_2,\sigma_2^2)$,
%
%
where we set $\mu_1 = 0.055, \mu_2 = 0.5, \sigma_1 = 0.05, \sigma_2 = 0.08, A_1=0.125, A_2=0.085$.

\paragraph{Quantitative results.}
Fig \ref{fig:latents} (d) shows the prior (blue) matches the posterior (red). Table \ref{tab:baselines} (fourth row) shows that our method achieves the lowest MSE among all baselines and a reconstruction accuracy comparable to the temporal VAE baseline.

\paragraph{Intervention results.}
Now, if we want to visualise how the pulsar would look like at a different frequency, we can replace the intensity profile prior with the intensity profile for that frequency. For example, the intensity profile for the Crab pulsar at 1418 MHz is shown in fig. \ref{fig:intensity-pulsar}, right, in black \cite{lyne_pulsars_2012} (fig. 9.2 of the cited paper). Again, we fit a Gaussian mixture model consisting of two Gaussians to this data (fig. \ref{fig:intensity-pulsar}, right, red) to obtain the intervention intensity prior as $I_{int}(t)=A_3\mathcal{N}(t;\mu_3,\sigma_3^2)+A_4\mathcal{N}(t;\mu_4,\sigma_4^2)$, 
%
%
where $\mu_3 = 0.195, \mu_4 = 0.6, \sigma_3 = 0.007, \sigma_4 = 0.008, A_3 = 0.0193, A_4 = 0.006$. The intervention curve is visualised in fig. \ref{fig:latents} (d), green, and is decoded to obtain a corresponding series of intervention images for the new frequency (fig. \ref{fig:video} (d), bottom).

\section{Conclusion}
In this work we presented a novel method based on a VAE with an ANM prior to learn physical variables from videos, that can be used to imagine novel (counterfactual) physical scenarios. After learning variables related by a specified physical prior for each video, we were then able to intervene on the prior to synthesize videos of the same physics experiment under different conditions, such as a pendulum with added air drag or a mass on a spring with modified object mass and frequency. To achieve this we proposed a novel method to approximate the non-linear ANM prior using a Gaussian Mixture Model where the shape of each Gaussian depends on the curvature of the prior. To efficiently optimise the VAE ELBO term we derived a numerically stable and parallelizable Monte Carlo estimate of the KL divergence between the posterior and prior GMMs. We have shown that the method is successful in learning the correct variables and generating realistic counterfactual videos for different real physics videos, and we have also proven theoretically that it recovers the correct variables. In the future we aim to expand the method to a broader class of videos and relax the assumption that the priors have to be fully specified.

\paragraph{Acknowledgements.} The authors acknowledge the generous support of the Royal Academy of Engineering (RF\textbackslash 201819\textbackslash 18\textbackslash 163).

\newpage
\bibliographystyle{plain}
\bibliography{egbib}

\end{document}